\documentclass{bmvc2k}

\usepackage{graphicx}
\usepackage{microtype}
\usepackage{amsmath}
\usepackage{amssymb}
\usepackage{algorithm}
\usepackage{xcolor}
\usepackage{bm}

\usepackage{multicol}
\usepackage{multirow}
\usepackage{dsfont}
\usepackage{mathtools}
\usepackage{enumitem}
\usepackage{colortbl}
\usepackage{arydshln}
\definecolor{headercolor}{gray}{0.93}
\definecolor{Gray}{gray}{0.9}
\newcolumntype{a}{>{\columncolor{headercolor}}c}
\usepackage{booktabs,multirow,multicol,tabularx}

\newcolumntype{Y}{>{\centering\arraybackslash}X}

\title{Stacked Temporal Attention:\\ Improving First-person Action Recognition by Emphasizing Discriminative Clips}

\addauthor{Lijin Yang}{yang-lj@iis.u-tokyo.ac.jp}{1}
\addauthor{Yifei Huang}{hyf@iis.u-tokyo.ac.jp}{1}
\addauthor{Yusuke Sugano}{sugano@iis.u-tokyo.ac.jp}{1}
\addauthor{Yoichi Sato}{ysato@iis.u-tokyo.ac.jp}{1}

\addinstitution{
 Institute of Industrial Science\\
 The University of Tokyo\\
 Tokyo, Japan
}

\runninghead{YANG ET AL.}{Stacked Temporal Attention}

\def\eg{\emph{e.g}\bmvaOneDot}

\def\etal{\emph{et al}\bmvaOneDot}
\newcommand{\model}{STAM\xspace}
\newcommand{\revise}[1]{\textcolor{black}{{#1}}}

\begin{document}

\maketitle

\begin{abstract}
First-person action recognition is a challenging task in video understanding. Because of strong ego-motion and a limited field of view, many backgrounds or noisy frames in a first-person video can distract an action recognition model during its learning process. To encode more discriminative features, the model needs to have the ability to focus on the most relevant part of the video for action recognition. 
Previous works explored to address this problem by applying temporal attention but failed to consider the global context of the full video, which is critical for determining the relatively significant parts.
In this work, we propose a simple yet effective Stacked Temporal Attention Module (\model) to compute temporal attention based on the global knowledge across clips for emphasizing the most discriminative features. We achieve this by stacking multiple self-attention layers. Instead of naive stacking, which is experimentally proven to be ineffective, we carefully design the input to each self-attention layer so that both local and global context of the video is considered during generating the temporal attention weights. Experiments demonstrate that our proposed \model can be built on top of most existing backbones and boost the performance in various datasets.
\let\thefootnote\relax\footnotetext{\hspace{-1.9em}Corresponding author: Yifei Huang}
\end{abstract}


\section{Introduction}\label{sec:introduction}

Video action recognition is one of the most fundamental problems in computer vision. Most research focus on recognizing human actions from videos taken by fixed cameras~\cite{two_stream,i3d,feichtenhofer2019slowfast} (third-person videos). Different from these third-person videos, first-person videos taken by wearable cameras record human behaviors from a different perspective, with a wide range of applications such as assisting human-computer interaction. Unfortunately, the performance of action recognition methods in first-person videos are still not comparable to that in third-person videos~\cite{sudhakaran2019lsta}. 

\begin{figure*}[]
    \centering
    \includegraphics[width=\linewidth]{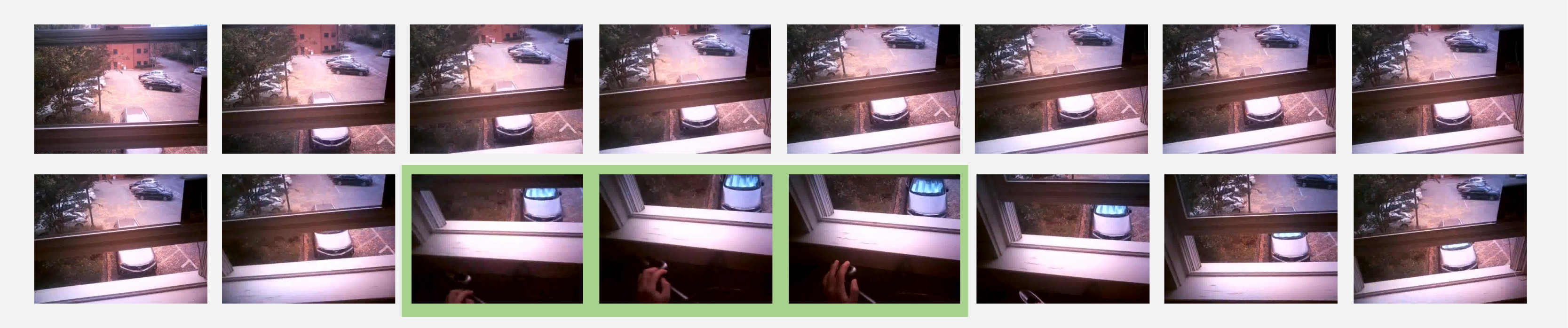}
    \vspace{-0.5cm}
    \caption{Example of action ``turn off faucet'' in the EGTEA dataset~\cite{li2018eye}. Only the frames with green boundaries provide enough information for determining the action. In all other frames, the action happens outside the visible field-of-view.}\vspace{-0.3cm}
    \label{fig:concept}
\end{figure*}

One of the major challenges for first-person action recognition is that the unique field of view makes actions sometimes happen outside the video viewing range (Figure~\ref{fig:concept}). Since the action is not always observable in all frames of the video sequence, the methods that first sparsely sample clips from the video and then compute the clip consensus via pooling~\cite{wang2016temporal,lin2019tsm}, which are proved to be good practices for creating more discriminative features in third-person videos, cannot achieve as good performance in first-person videos.

Another major challenge for first-person action recognition is that the videos are often accompanied by huge ego-motion caused by the sharp movements of the camera wearer (\eg turning head), which complicates the encoding of motion~\cite{sudhakaran2019lsta}. With current clip sampling techniques, some of the clips may contain purely background information that harms the recognition, even with an enlarged temporal receptive field. 

One way to address the two challenges jointly is to apply temporal attention to the clips, emphasizing the discriminative clips while suppressing the noisy ones. Several previous works have explored this direction. For example, Pei \textit{et al.}~\cite{pei2017temporal} propose to use gated LSTM for deciding the weight of each input frame, and Girdhar \textit{et al.}~\cite{girdhar2019video} leverage transformers~\cite{vaswani2017attention} to compute spatio-temporal attention on humans in the video. 
However, either using LSTM or transformers, the temporal information is only formulated by computing clip-wise relationships. It is still difficult for the model to decide the best choice of attention without a global understanding of the whole input.
Intuitively thinking, global knowledge of all the local clips can help to identify which clip is the most discriminative one and suppress the less important clip features.

In this work, building upon this intuition we propose a Stacked Temporal Attention Module (\model) that leverages both local information from individual clips and global information from the whole video to generate temporal attention. We find that this could be done by simply stacking self-attention~\cite{vaswani2017attention} layers with global information as the query vectors. 
It is worthwhile to note that our \model can be built on top of most existing action recognition backbones. Experiments on multiple first-person datasets demonstrate that by adding \model on top, the action recognition performance could be improved. Further studies on HMDB51 dataset~\cite{kuehne2011hmdb} show that our method can even generalize to third-person datasets.

Our major contributions are: (1) We propose a simple yet effective Stacked Temporal Attention Module (\model) that could be built on top of most existing backbones for improving first-person action recognition. (2) Experiments on multiple datasets demonstrate that our proposed module can improve the performance of various backbone models.
\section{Related Work}\label{sec:related_work}
\paragraph{First-person action recognition}
Action recognition is one of the most focused research areas in computer vision, and first-person action recognition is also extensively studied. Earlier works design different hand-crafted features~\cite{fathi2012learning,li2015delving} to represent the rich spatio-temporal information in the video sequence. \revise{With the development of deep learning, remarkable progress has been made recently~\cite{ma2016going,sudhakaran2018attention,lu2019learning,kapidis2019multitask,huang2020improving,huang2020ego,li2021ego,girdhar2021anticipative}.} Since temporal information is essential for recognizing the actions in the videos, a line of research~\cite{sudhakaran2019lsta, li2018videolstm} uses recurrent networks for sequential modeling. However, one major drawback of using recurrent models is that the span of attention is limited~\cite{singh2016multi}, which prevents these models to perform well when the input video is long. Some previous works also tried to incorporate unique first-person cues such as hand and gaze~\cite{singh2016first,lu2019learning,zuo2018gaze,huang2020mutual,huang2018predicting}, or use multiple modalities~\cite{kazakos2019epic} for improving first-person action recognition.
Some researchers try to input multiple frames and use spatio-temporal convolution networks for aggregating features both spatially and temporally~\cite{li2018eye,wang2020symbiotic}. 
One problem of 3D convolution networks is that the limited temporal receptive field results in a model that lacks long-term temporal knowledge of the input.
In this work, we use these models as backbones and build our Stacked Temporal Attention Module on top for further improving their performances.

\paragraph{Temporal attention for action recognition}
Previous works tried different temporal attention methods to aggregate the temporal features for longer-term video understanding and have shown promising results in action recognition of third-person videos.
One line of research use temporal aggregation methods to implicitly apply different weights on each feature~\cite{zhou2018temporal,he2019stnet,wu2019long}
For example, TSN~\cite{zhou2018temporal} sparsely samples frames from videos and assign uniform temporal attention by average pooling to combine features of each frame. TLE~\cite{diba2017deep} designed a temporal linear encoding layer for generating weight for each temporal input.

Another line of research explicitly computes a temporal attention score for each feature and uses a weighted average for the aggregation. As for previous works, TAGM~\cite{pei2017temporal} modified the LSTM block and sequentially determine a weight for each input frame. One major drawback is that LSTM cannot access global information, thus the attention produced by only local features is sub-optimal.
Similar to our method, CatNet~\cite{wang2020cascade} also uses global information to aggregate the clip features. However, they simply concatenate the local and global features and use several fully connected layers for predicting the weights. The attention produced by these fully connected layers is used as temporal attention weights.

In this work, we design a simple yet effective module that could be built on top of most action recognition models for computing temporal attention. We find that simply by stacking self-attention layers~\cite{vaswani2017attention} with global feature vectors as queries, we can improve recognition performance on multiple first-person datasets.

\begin{figure*}
    \centering
    \includegraphics[width=\linewidth]{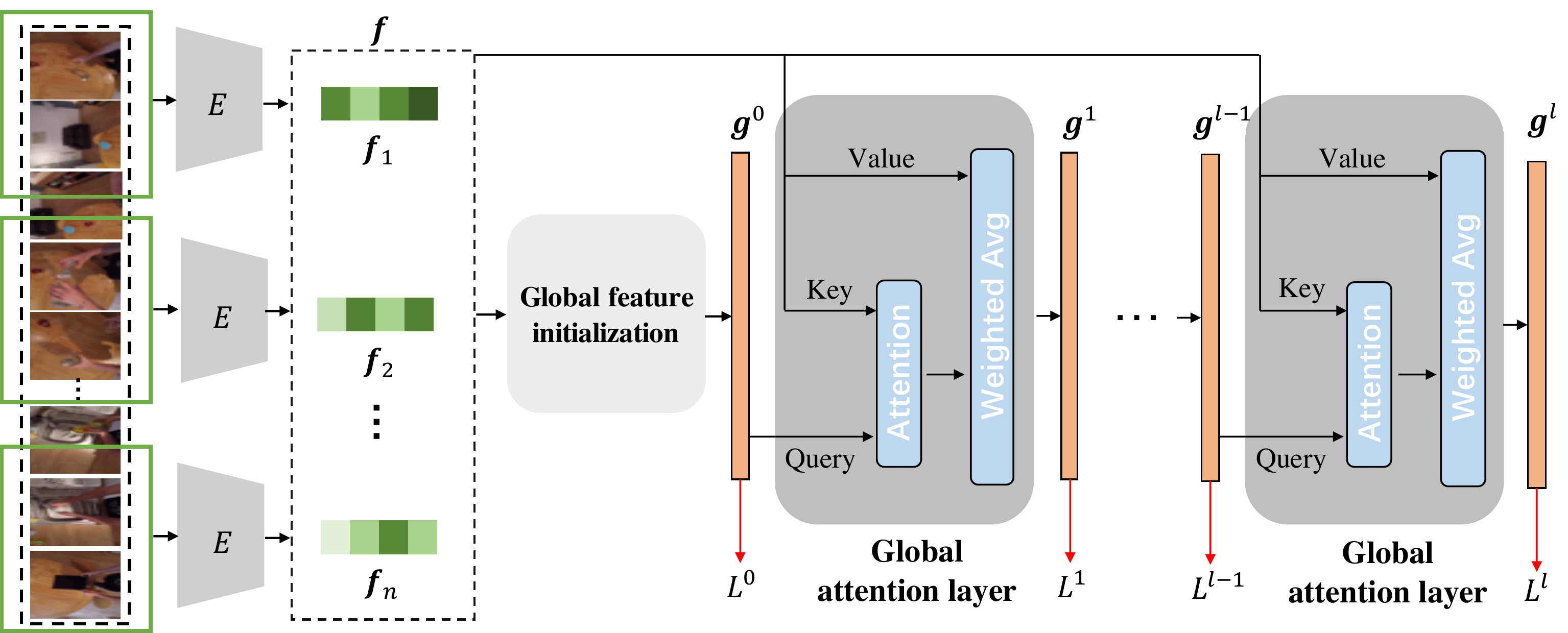}
    \vspace{-0.3cm}
    \caption{Overview of our proposed \model. Using the features from the backbone, a global feature $\bm{g}^0$ is first initialized. This global feature serves as an aggregated global information of all input clips and is used for the following global attention layers. The stack of global attention layers progressively refines the attention weight as well as the global feature. Classification loss is applied to the output of each layer. }\vspace{-0.3cm}
    \label{fig:model}
\end{figure*}
\section{Approach}\label{sec:approach}
In this section, we describe our proposed Stacked Temporal Attention Module (\model). Different from existing methods for temporal attention~\cite{wang2016temporal,pei2017temporal,zhou2018temporal}, we consider the global information across clips, and use the relation between a clip and the global representation to determine the significance of each clip.
An overview of our model architecture is depicted in Figure~\ref{fig:model}.
\vspace{-0.3cm}
\paragraph{Backbone encoder}
Our \model can cooperate with most of the existing backbones that use a neural network and back-propagation for training. Here we showcase a backbone encoder $E$ that uses 3D convolution and takes as input a small video clip with a few video frames. Given a trimmed video containing an action, we follow the practice of TSN~\cite{wang2016temporal} to split the input into $N$ segments. From each segment we extract a clip of 16 consecutive frames as inputs to the backbone network. The backbone would encode the clips into clip features $\bm{f}_1, ...,\bm{f}_N$. These clip features are used as inputs to our \model.

\subsection{Stacked Temporal Attention Module}
With the features encoded by the backbone, our \model finds attention for temporally aggregating the features using both local and global information. As shown in Figure~\ref{fig:model}, the \model is a stack of several global attention layers. The input to the first global attention layer is the global feature vector $\bm{g}^0$ initialized by the local feature vectors $\bm{f}$. We explore multiple initialization methods in this work. In the $l$-th global attention layer, an attention weight vector $\bm{a}^{l}$ is computed using the relation between the input global feature $\bm{g}^{l-1}$ and each local feature $\bm{f}$. This attention weight is used to generate a new global vector $\bm{g}^{l}$ as the output of this layer. For simplicity, we omit all the classifiers and normalization layers in this section.

\subsubsection{Global Attention Layer}
As in Figure~\ref{fig:model}, the global attention layer takes the encoded clip features and the global feature $\bm{g}$ as inputs. Intuitively, knowing the global context can help to determine the importance of each local clip. Thus, the relationship between the global feature and each of the clip features should be considered to emphasize discriminative clips for getting better temporal attention. This could be neatly done taking advantage of the self-attention operation of the 
Transformers \cite{vaswani2017attention}. Take the first global attention layer as an example, the query vector is acquired using the global feature $\bm{g}_0$, while the key and value vectors are from each local features $\bm{f}$:
\vspace{-0.1cm}
\begin{gather}
    \bm{q}_i = W_q\bm{g}^0, \qquad \bm{k}_i = W_k\bm{f}_i, \qquad \bm{v}_i = W_v\bm{f}_i;\quad i \in [1,N]
\end{gather}
where $W_q, W_k, W_v$ are the weight matrices for projecting the features to query $\bm{q}$, key $\bm{k}$ and value $\bm{v}$, respectively. The queries and keys are vectors with hidden dimension $d$, and the value vectors have the same dimension as the backbone features. We then compute a weight vector $a^{1}_i$ for the $i$-th clip using a modified scaled dot-product operation and softmax operation:
\vspace{-0.2cm}
\begin{equation}
    c_i = \frac{1}{N\sqrt{d}} \sum_{j=1}^N \bm{q}_i \cdot {\bm{k}_j}, \qquad
    a^{1}_i = \frac{e^{c_i}}{\Sigma_j e^{c^j}}
\end{equation}
The generated weight vectors are used as temporal attention weights:
\vspace{-0.2cm}
\begin{equation}
    \bm{g}^{1} = \sum_{i=1}^N a^{1}_i \bm{v}_i
\end{equation}
$\bm{g}^1$ is the output of the first global attention layer and will also be the input to the next layer. We add a classifier on top of $\bm{g}^{1}$ to recognize actions and omit it in the equations for simplicity.

Stacking several modules sequentially has shown significant improvements in many tasks like human pose estimation~\cite{newell2016stacked} and action segmentation~\cite{farha2019ms}. By composing several models sequentially, each model can perform an incremental refinement to the previous output. Inspired by the previous success, in this work we propose to stack multiple global attention layers for better refining the temporal attention and the global features.

Denote the operation of the first global attention layer as $\bm{g}^1 = \mathcal{G}^1(\bm{g}^0, \bm{f}_1, ..., \bm{f}_N)$, the k-th global attention layer can be represented accordingly as follows:
\begin{equation}
    \bm{g}^k = \mathcal{G}^k(\bm{g}_{k-1}, \bm{f}_1, ..., \bm{f}_N)
\end{equation}
We show in our experiments that as the global feature refines, the attention scores also change, giving more focus on the most critical clip from a global perspective. 

\subsubsection{Global feature initialization}
Given the encoded clip features $\bm{f}_1, \cdots, \bm{f}_N$ as inputs, the global feature vector $\bm{g}^0$ can be initialized in multiple ways. We experiment with the following choices while others are possible:

\vspace{0.1cm}
\noindent\textbf{Pooling:} One of the most straightforward ways is to do pooling on all clip features to get the global feature: $\bm{g}^0 = pool(\bm{f}_1, \cdots, \bm{f}_N)$ We experiment with both max-pooling and average-pooling in the experiment section.

\vspace{0.1cm}
\noindent\textbf{Recurrent networks:} It is also feasible to use recurrent neural network to process all clip features, and generate a global aggregated feature. We choose bi-directional GRU as an example: $\bm{g}^0 = GRU(\bm{f}_1, \cdots, \bm{f}_N)$.

\vspace{0.1cm}
\noindent\textbf{Temporal convolution:} Apply a 1D temporal convolution layer to fuse all the clip features: $\bm{g}^0 = Conv(\bm{f}_1, \cdots, \bm{f}_N)$.

\vspace{0.1cm}
\noindent\textbf{Light-weight CNN:} Inspired by previous works ~\cite{wu2019adaframe,korbar2019scsampler}, we also tried to use a light-weight CNN for extracting the global feature. The lightweight CNN is a small 2D CNN~\cite{howard2017mobilenets} that takes one frame from each input clip as inputs. The output features of all inputs are averaged to form the global feature.

\vspace{0.1cm}
\noindent\textbf{Self-attention:} We use the self-attention operation of Transformers~\cite{vaswani2017attention} to compute the pairwise relationship between input clips, and use generated weight to form the global feature. 

\subsection{Model training}
To incrementally refine global features of each layer, we add a classifier with cross-entropy loss on each attention layer during training the model. The final loss function is a combination of the loss from both the global feature initialization step and $M$ global attention layers:
\begin{equation}
    L = \sum_{i=0}^{M}\lambda_i L^i
\end{equation}
Here $\lambda_0, \cdots ,\lambda_M$ is a set of model hyper-parameters to determine the contribution of the different losses.

\section{Experiments and Results}\label{sec:experiment}
\revise{We conduct experiments on two publicly available first-person datasets: EGTEA~\cite{li2018eye,li2021eye} and EPIC-Kitchens-100~\cite{damen2020rescaling} (EPIC-100).}  EGTEA is an egocentric video dataset containing 29 hours of meal preparing actions performed in a kitchen by 32 subjects. Fine-grained
annotations of 106 action classes are provided. Following~\cite{sudhakaran2019lsta}, we report the average of three splits of the dataset. EPIC-100 is the currently largest dataset of first-person videos, with 100 hours of videos recorded at different kitchens by different subjects. Fine-grained annotations of 97 verbs and 300 nouns are provided. We follow the official protocol and report the performance of verb, noun and the combined actions. For all datasets, we report the action recognition accuracy as the evaluation metric.

\revise{\noindent\textbf{Implementation Details.} We use PyTorch~\cite{pytorch} for all the implementation. All backbones are pretrained on Kinetics dataset~\cite{carreira2017quo}. We train \model together with pretrained backbone in the end-to-end manner using the Adam optimizer~\cite{kingma2014adam} with initial learning rate 1e-4 for 40 epochs. As for the loss weight $\lambda$ we empirically set all the $\lambda$s to 1.  }

\revise{The hidden dim $d$ is set as 512 in all experiments. 
For the 2D backbone TSM, we uniformly sample 16 frames as the input. The output temporal dimension is the same as clip number for both 2D and 3D backbones. For the number of global attention layer $M$, we empirically set different values to cooperate with different backbone encoders. More details about implementation can be found in the supplementary material.}

\begin{table}[h]
    \begin{minipage}[]{.48\textwidth}
        \vspace{0pt}
        \centering
        \scalebox{0.76}{
        \begin{tabular}{p{3.2cm}<{\centering}p{0.9cm}<{\centering}p{0.8cm}<{\centering}p{0.8cm}<{\centering}}
          \toprule
          \multirow{2}{*}{\textbf{Global feature options}}&
          \multicolumn{3}{c}{\textbf{Num of global att layers}}\cr
          \cmidrule(lr){2-4}&1&2&3 \cr 
          \midrule
          \rowcolor{Gray}
          \revise{Vanilla stacking} & 65.6 & 65.5 & 65.5 \cr
          Max pooling  &65.8&65.9&66.3\cr
          Avg pooling  &65.9&66.3&66.7\cr
          Bi-GRU  &65.8&66.3&65.9\cr
          1D-conv &65.9&66.1&65.9\cr
          Self-att &\textbf{66.2}&\textbf{66.5}&66.9\cr
          Light-weight CNN&64.7&66.0&\textbf{67.1}\cr
          \bottomrule
        \end{tabular}}
        \vspace{0.1cm}
        \caption{Different design options for the global feature. Experiments are conducted on the EGTEA dataset split1 using I3D as backbone and 9 clips as inputs. Dark row indicates no global feature is used.}
        \label{tab:options}
    \end{minipage}
    \hspace{0.1cm}
    \begin{minipage}[]{.48\textwidth}
        \vspace{0pt}
        \centering
        \scalebox{0.76}{
        \begin{tabular}{ccccc}
          \toprule
          \multirow{2}{*}{\textbf{Method}} & \multirow{2}{*}{\textbf{EGTEA}} & \multicolumn{3}{c}{\textbf{EPIC-100}} \cr
          \cmidrule(lr){3-5}&&Verb&Noun&Action \cr 
          \midrule
          Max pooling &63.6&61.8&45.6&33.7\cr
          Avg pooling &63.0&63.2&48.0&36.2\cr
          Bi-GRU &62.5&62.3&45.1&33.7\cr
          1D-conv&62.6&63.4&46.3&35.2\cr
          Self-att&65.5&62.9&48.6&36.8\cr
          \\[-0.8em]
          \hdashline
          \\[-0.9em]
          TAGM~\cite{pei2017temporal}&65.4&63.6&45.7&34.3\cr
          CatNet~\cite{wang2020cascade} &63.2&61.5&48.7&35.8\cr
          TLE~\cite{diba2017deep}&63.4&63.6&45.4&34.3\cr
          TRN~\cite{zhou2018temporal} & 64.0 & 64.6 & 47.9&36.4\cr
          MTRN~\cite{zhou2018temporal} & 64.1 &\textbf{64.8}&47.6&36.6\cr
          \revise{GSTA~\cite{suin2021gated}} & 63.8 & 62.2 & 49.1 & 36.5\cr
          \revise{STA~\cite{li2020spatio}} & 63.8 & 62.6 & 48.6 & 36.6\cr
          \revise{CTA~\cite{wharton2021coarse}} & 62.6 & 62.3 & 47.2 & 35.4\cr
          
          \\[-0.8em]
          \hdashline
          \\[-0.9em]
          Our \model &\textbf{66.9}&64.4 & \textbf{50.3} & \textbf{38.0}\cr
          \bottomrule
        \end{tabular}}
        \vspace{0.1cm}
        \caption{Comparison of different feature aggregation methods on EGTEA dataset split1 and EPIC-100 dataset.}\vspace{-0.5cm}
        \label{tab:agg}
    \end{minipage}
\end{table}

\subsection{Initialization options of global feature}
\label{sec:options}
One core component in our \model is the use of global feature $\bm{g}$ for temporal attention computation. It is possible to utilize multiple alternatives as the initialization of global feature as described in Section~\ref{sec:approach}. 
Table~\ref{tab:options} shows the performance of our \model when changed with different alternatives of global feature $\bm{g}$.
\revise{We further add directly stacking self-attention layers as the vanilla stacking baseline. Note that vanilla stacking baseline does not include global feature initialization, we use one more self-attention layer instead to keep layer number fair with our \model using self-att initialization.}
From Table~\ref{tab:options}, we can see that our proposed \model is not too sensitive about the initialization option of global feature. 
We empirically find that using self-attention for global feature initialization can get the most stable performance in different settings (\eg different number of input clips). 
We think one possible reason is that self-attention alone can be a strong baseline for action recognition (validated in Table~\ref{tab:agg}). Thus, we use this option in all of the following experiments unless otherwise stated.

 \revise{It is worthwhile to note that directly stacking self-attention layers can hardly improve model performance, which shows the same trend as in \cite{neimark2021video}. However, our \model with all initialization options can benefit from stacking more than one layer. We believe this is because although the vanilla stacking baseline uses updated local feature as the key and query for the next layer, it still cannot access global information of all other local features at the same time. Also, simultaneously updating the key and query can lead to accumulated errors, which may even make performance worse when increasing layers. }
 
 \revise{
 Compared with vanilla stacking, our proposed method leverages the global feature and only refines the query when stacking, which enables the model to better determine the relatively significant clips as well as benefit from stacking. Therefore, our \model(self-attention for global feature initialization) can clearly surpass vanilla stacking baseline under fair comparison. This suggests the usefulness of \model for generating temporal attention, and shows potential applications in other tasks that also deal with temporal information, \eg, action detection.}

\subsection{Comparison of temporal attention methods}
\label{sec:agg}
For evaluation of our proposed temporal attention module, we compare our \model with some baselines and several state-of-the-art temporal attention methods. For EGTEA dataset we utilize I3D backbone and use 9 clips with 16 frames for all methods. For EPIC-100 dataset we use TSM backbone with 16 frames as inputs. We specifically compare with the following clip feature aggregation methods:\vspace{-0.1cm}

\begin{itemize}
    \item Global feature initialization baselines without the proposed \model. These baselines serve as simple straightforward ways for temporal aggregation. \vspace{-0.1cm}
    \item \textbf{TLE}~\cite{diba2017deep}, \textbf{TAGM}~\cite{pei2017temporal}, \textbf{CatNet}~\cite{wang2020cascade}, \textbf{TRN}~\cite{zhou2018temporal},
    \revise{
    \textbf{GSTA}~\cite{suin2021gated}, 
    \textbf{STA}~\cite{li2020spatio} and
    \textbf{CTA}~\cite{wharton2021coarse} }
    are previous works that uses various algorithms for temporal aggregation. \vspace{-0.1cm}
    \item Our \textbf{\model}. In this experiment, for EGTEA and EPIC-100 we use \model with three global attention layers and two global attention layers respectively. Studies on the influence of the number of global attention layers are placed in Section~\ref{sec:explore}. \vspace{-0.1cm}
\end{itemize}

Quantitative result comparison is shown in Table~\ref{tab:agg}. Using the same backbone and the same number of input clips, our \model significantly outperforms other feature aggregation methods. This strongly validates that our \model can effectively leverage the local and global information in the clip features to predict better temporal attention weights.

\subsection{Comparison with SOTA}
We conduct experiments to validate the improvement in performance that \model brought to the backbones and compare the performance with other state-of-the-art methods. We compare the following methods for first-person action recognition: Ego-RNN~\cite{sudhakaran2018attention}, LSTA~\cite{sudhakaran2019lsta} and SAP~\cite{wang2020symbiotic}.
We also add our proposed \model on 3 commonly used backbones: TSM~\cite{lin2019tsm}, I3D~\cite{i3d}, and 3D-ResNet-50 (R3D-50)~\cite{hara2018can}. For the 2D backbone TSM we input multiple frames, and for other backbones the inputs are multiple 16-frame video clips. 

\begin{table}[]
    \begin{minipage}[]{.4\textwidth}
        \vspace{0pt}
        \centering
        \scalebox{0.76}{
        \begin{tabular}{ccccc}
        \toprule
        \textbf{Method} & \textbf{Split1} & \textbf{Split2} & \textbf{Split3} & \textbf{Average} \cr
        \midrule
        Ego-RNN~\cite{sudhakaran2018attention}& 62.2 & 61.5 & 58.6 & 60.8 \cr
        LSTA*~\cite{sudhakaran2019lsta}& - & - & - &61.9\cr
        SAP~\cite{wang2020symbiotic}& 64.1 & 62.1 & 62.0 & 62.7  \cr
        \rowcolor{Gray}
        Lu \etal~\cite{lu2019learning} & 63.7 & 61.1 & 59.0 & 61.3 \cr
        \rowcolor{Gray}
        Min* \etal~\cite{min2021integrating} & 69.6 & -& -& - \cr
        TSM~\cite{lin2019tsm} & 63.8 & 61.8 & 60.2 & 61.9 \cr
        I3D~\cite{i3d}& 63.0 & 61.1 & 58.0 & 60.7\cr
        R3D-50~\cite{hara2018can}& 63.2 & 61.4 & 59.4 & 61.3 \cr
        \\[-0.8em]
        \hdashline
        \\[-0.9em]
        TSM + \model & 66.2 & \textbf{64.1} & \textbf{64.0} & \textbf{64.8} \cr
        I3D + \model & \textbf{66.9} & 63.8 & 62.2 & 64.3\cr
        R3D-50 + \model & 65.3 & 62.9 & 63.1 & 63.8\cr
        \bottomrule
        \end{tabular}}
        \vspace{0.1cm}
    \end{minipage}
    \hspace{0.1cm}
    \begin{minipage}[]{.7\textwidth}
        \vspace{0pt}
        \centering
        \scalebox{0.76}{
        \begin{tabular}{cccc}
        \toprule
        \multirow{2}{*}{\textbf{Method}} & \multicolumn{3}{c}{\textbf{Top-1}} \cr
        \cmidrule(lr){2-4}
        &Verb&Noun&Action \cr 
        \midrule
        TSM~\cite{lin2019tsm} & 63.2 & 48.0 & 36.2  \cr
        I3D~\cite{i3d}& 55.5 & 43.3 & 29.3  \cr
        R3D-50~\cite{hara2018can}& 56.7 & 45.3 & 31.5 \cr
        \\[-0.8em]
        \hdashline
        \\[-0.9em]
        TSM + \model & \textbf{64.3} & \textbf{50.3} & \textbf{38.0} \cr
        I3D + \model & 57.4 & 45.2 & 31.7 \cr
        R3D-50 + \model & 58.0 & 47.4 &  33.7\cr
        \bottomrule
        \end{tabular}}
        \vspace{0.1cm}
    \end{minipage}
\caption{Results on the EGTEA dataset (left) and EPIC-100 dataset (right). * indicates optical flow is used, dark rows indicate gaze information is used. All experiments of our \model only take RGB frames as inputs.}
\label{tab:baseline}
\end{table}

Table \ref{tab:baseline}\textcolor{red}{-left} and Table \ref{tab:baseline}\textcolor{red}{-right} show the result comparison on the EGTEA dataset and the EPIC-100 dataset, respectively. From both tables, we can see that the performance of all backbones can be improved when adding our proposed \model. On the EGTEA dataset, our method achieves the state-of-the-art performance among all the methods with RGB frames as inputs, and is even better than LSTA~\cite{sudhakaran2019lsta} which also uses optical flow. Only Min \etal~\cite{min2021integrating} outperforms our method, but they use optical flow and human gaze as additional information. Similarly, our method can boost the performance of all backbones on the EPIC-100 dataset. This performance increase is significant given the intrinsic difficulty of the EPIC-100 dataset (\eg non-scripted, unbalanced label, large variety of action).

\subsection{Exploration of \model}
\label{sec:explore}
For all temporal aggregation methods, the number of input clips (temporal receptive field) can have a huge influence on the final performance.
We explore the performance of our \model with different numbers of clips as inputs.

\begin{table*}
    \centering
    \scalebox{0.76}{
    \begin{tabular}{cccccccc}
    \toprule
    \multirow{2}{*}{\textbf{Clip Num}}&
    \multirow{2}{*}{\textbf{Avg}}&
    \multicolumn{5}{c}{\textbf{Num of global att layers}}&
    \multirow{2}{*}{\textbf{Gain}}\cr
    \cmidrule(lr){3-7}&&0&1&2&3&4& \cr 
    \midrule
    3 & 59.1&61.1&61.7&62.2&\textbf{63.5}&62.5&+4.4\cr
    6 & 62.3&64.3&65.1&\textbf{65.6}&65.1&64.9&+3.3\cr
    9 & 63.0&65.5&66.2&66.5&\textbf{66.9}&66.7&+3.9\cr
    \bottomrule
    \end{tabular}}
    \vspace{0.1cm}
    \caption{Analysis of the influence of input clip (temporal receptive field) and the number of stacked global attention layers. Experiments are done using I3D backbone on the EGTEA dataset split1.}
    \label{tab:coop1}\vspace{-0.3cm}
\end{table*}

Table \ref{tab:coop1} shows the influence of different numbers of attention layers used in our \model. We experiment with 3, 6 and 9 input clips. In both tables, \textit{Avg} indicates direct averaging of all clip features, which serves as a baseline for understanding the usefulness of temporal attention. \textit{Gain} emphasize the maximum performance gain with respect to the \textit{Avg} method. Usually, since using more clips brings richer information, the recognition accuracy tends to become higher.

From the table, we can see that with the help of \model, the performance gain of a small temporal receptive field (3 clips) is more obvious. This from one side proves that our \model can learn to highlight the most important clip feature.
The table also suggests that stacking two layers (with 6 clips as inputs) or three layers (with 3 or 9 clips as inputs) of global attention works the best, and stacking four layers saturates the performance. 

\begin{figure*}
    \centering
    \includegraphics[width=1.0\linewidth]{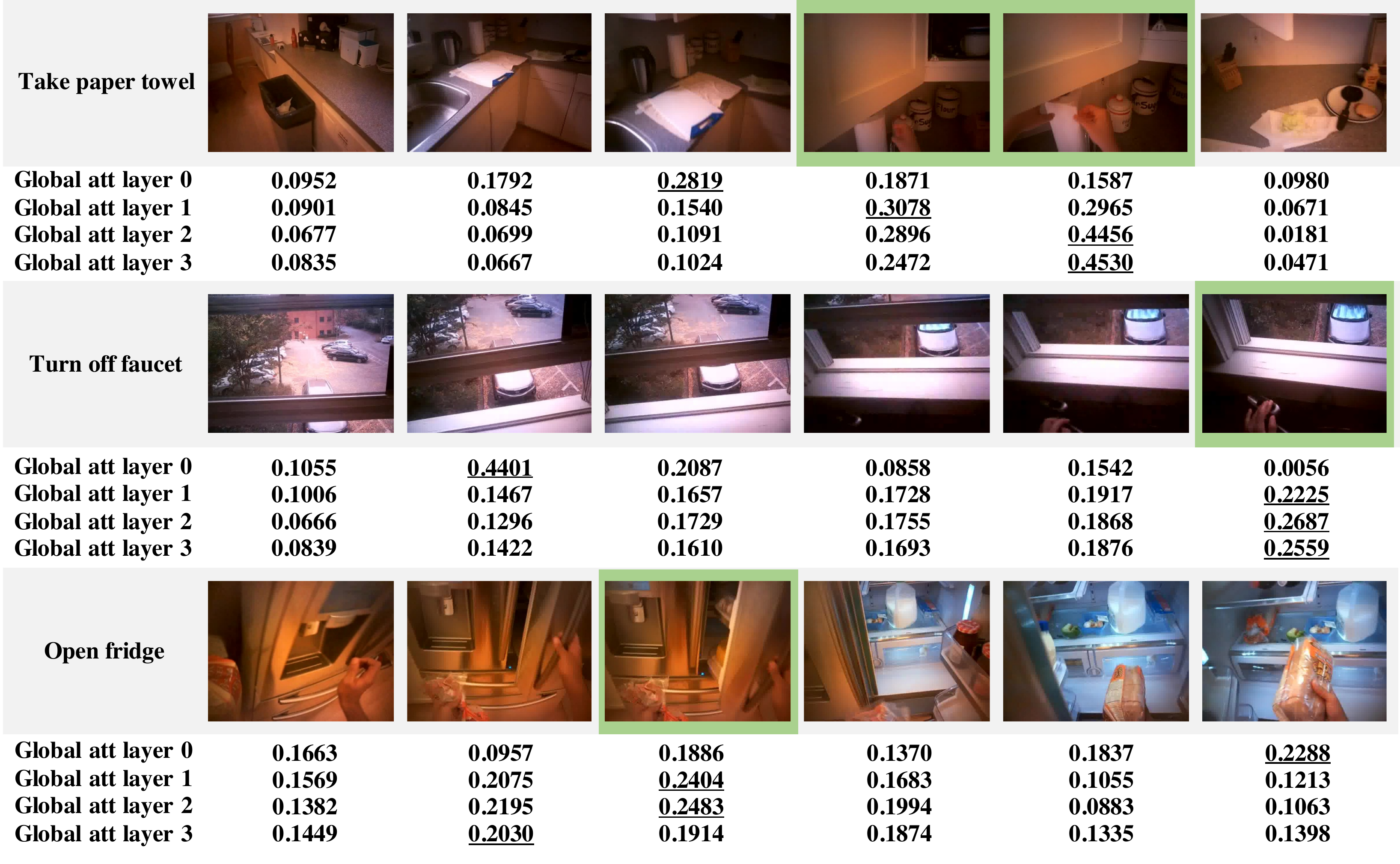}
    \vspace{-0.5cm}
    \caption{Visualization of temporal attention score of a few samples of the EGTEA dataset when stacking different numbers of global attention layers. The highest temporal attention score in each layer is underlined. In the case of global attention layer 0, initialization using the self-attention method is used. The frame with green background indicates that the backbone model can predict the correct action class with this clip alone as input.}
    \vspace{-0.3cm}
    \label{fig:vis}
\end{figure*}

Figure~\ref{fig:vis} shows visualizations of several sample actions in the EGTEA dataset together with the temporal attention for each clip. We use one frame to represent each 16-frame clip in this Figure, showing the model performance when using 6 clips as inputs. In each global attention layer, the highest temporal attention score is marked with an underline. Frames with green boundaries indicate that only with the corresponding clip as input, the network can predict the correct action class. For all the samples, with the stack of global attention layer, we can see the temporal attention gradually shifts, giving the correct clip higher weights. For example, in the action ``take paper towel'', the initialization of global layer using self-attention (global layer 0) gives more attention to a background clip. This is reasonable since with self-attention on each clip's feature, the network tends to assign high values to clips that have similar features. With the help of the global attention layer, the highest temporal attention goes to the 4th and 5th clip, which contains sufficient information for correct action recognition and therefore should be emphasized. This strongly supports that the use of global feature is important. After stacking more global attention layers, the temporal attention is refined to give more focus on the correct clips. Other samples of ``turn off faucet'' and ``open fridge'' show similar performance. This proves that the proposed stacking of the global attention layers is effective in refining the temporal attention. In this Figure, the temporal attention score becomes optimal when stacking 2 global attention layers, and saturates if we stack more than three layers, showing a similar trend with Table~\ref{tab:coop1}.

\subsection{Experiments on third-person dataset}
Furthermore, to see how the proposed \model performs on traditional third-person datasets, we add an experiment using the HMDB51 dataset~\cite{kuehne2011hmdb}.
HMDB51 is a widely used human motion dataset collected from YouTube that contains 6849 clips divided into 51 action categories. The average duration of each action is about 3s. Following the convention, we report the average result of three train/test splits.

With our proposed \model, the performance of TSM (69.4 $\rightarrow$70.3), I3D (73.1$\rightarrow$76.1) and 3D ResNet (67.8$\rightarrow$70.5) backbones are all improved. The improvement may not be as significant as in first-person videos, but it strongly proves the effectiveness of the proposed \model, and demonstrates its generalization ability.
Due to the page limit, the complete table of quantitative result comparison can be found in the supplementary material.

\section{Conclusion}
In this paper, we propose a Stacked Temporal Attention Module (\model) that generates temporal attention to better aggregate clip features for first-person action recognition. Our \model can be built on top of most existing models to improve their action recognition performance. Experiments on EPIC-100 and EGTEA datasets demonstrate that our \model is capable of boosting the performance of multiple backbones. Experiments on the HMDB51 dataset suggests that our \model also works on third-person videos.
As for our future work, we plan to explore other design choices for constructing the global feature, and leveraging the stacked temporal attention on other video tasks.

\section*{Acknowledgement}
This work was supported by JST AIP Acceleration Research Grant Number JPMJCR20U1 and JSPS KAKENHI Grant Number JP20H04205, Japan.

\bibliography{bib}

\end{document}